\documentclass[conference,a4paper]{IEEEtran}
\IEEEoverridecommandlockouts

\usepackage[hidelinks]{hyperref}
\usepackage[cmex10]{amsmath}
\usepackage{amssymb,amsfonts}
\usepackage{dblfloatfix}

\usepackage[ruled,vlined]{algorithm2e}
\usepackage{graphicx}
\graphicspath{{Figures/PDF/}{Figures/PNG/}}
\usepackage{multirow}
\usepackage{booktabs}
\usepackage{siunitx}
\usepackage[numbers,compress]{natbib}
\usepackage{texnames}
\usepackage{bm,bbm}
\usepackage{orcidlink}
\usepackage{tabularx}
\begin{document}
\title{\uppercase{Foundation AI Models for Aerosol Optical Depth Estimation from PACE Satellite Data}
\thanks{Copyright 2026 IEEE. Published in the 2026 IEEE International Geoscience and Remote Sensing Symposium (IGARSS 2026), scheduled for 9 - 14 August 2026 in Washington, D.C.. Personal use of this material is permitted. However, permission to reprint/republish this material for advertising or promotional purposes or for creating new collective works for resale or redistribution to servers or lists, or to reuse any copyrighted component of this work in other works, must be obtained from the IEEE. Contact: Manager, Copyrights and Permissions / IEEE Service Center / 445 Hoes Lane / P.O. Box 1331 / Piscataway, NJ 08855-1331, USA. Telephone: + Intl. 908-562-3966.}
}

\author{	
    \IEEEauthorblockN{Zahid Hassan Tushar\orcidlink{0000-0002-8231-6767},
    and Sanjay Purushotham}
    \IEEEauthorblockA{\textit{University of Maryland, Baltimore County, Maryland, USA.}\\
        Emails: \{ztushar1, psanjay\}@umbc.edu}
}

\maketitle
\begin{abstract}
Aerosol Optical Depth (AOD) retrieval is essential for Earth observation, supporting applications from air quality monitoring to climate studies. Conventional physics-based AOD retrieval methods formulate the problem as a pixel-wise inversion, relying on radiative transfer modeling, memory-intensive look-up tables, and auxiliary meteorological data. While recent data-driven approaches have shown promise, many fail to exploit the spatial–spectral coherence of hyperspectral imagery, leading to spatially inconsistent and noise-sensitive retrievals.
We present the first study exploring \textit{Foundation AI models for AOD retrieval} and propose \textit{ViTCG, a Vision Transformer with Channel-wise Grouping} -based spatial regression framework that reduces retrieval bias and error. ViTCG uses hyperspectral top-of-atmosphere radiance as input and jointly models spatial context and spectral information. Validation with PACE radiance observations demonstrates a 62\% reduction in mean squared error compared to state-of-the-art foundation models, including \textit{Prithvi}, and produces spatially coherent AOD fields.    
\end{abstract}


\section{Introduction}

Aerosol Optical Depth (AOD) is a fundamental Earth observation parameter that quantifies column-integrated aerosol loading and plays a central role in radiative forcing assessments and air quality monitoring~\cite{kaufman2002satellite,remer2005modis}. Satellite-derived AOD products are widely used to estimate surface-level particulate matter ($PM_{2.5}$), particularly in regions lacking dense ground-based observations such as AERONET~\cite{holben1998aeronet}, by integrating AOD with chemical transport models and meteorological data~\cite{lee2012use,di2019ensemble}. High-resolution and spatially consistent AOD retrievals are therefore critical for climate studies, aerosol transport analysis, and the monitoring of aerosol-related hazards, including wildfire smoke and dust events~\cite{bellouin2020bounding,huff2021tracking}.

Conventional satellite AOD retrieval algorithms are typically formulated as pixel-wise inverse problems. Operational approaches for sensors such as MODIS, VIIRS, MISR, and geostationary imagers rely on radiative transfer modeling, look-up tables, or temporal compositing techniques to infer AOD from multispectral observations~\cite{mei2012retrieval,remer2019retrieving}. While physically interpretable, these methods process pixels independently and do not explicitly exploit spatial coherence, leading to increased uncertainty over heterogeneous land surfaces and complex terrain where the inversion problem is weakly constrained~\cite{she2020himawari,chen2022himawari}.

Recent studies have explored machine learning and deep learning methods to improve AOD retrieval accuracy by learning nonlinear relationships between spectral observations and aerosol loading~\cite{yeom2021estimation,kang2022improved}. Although approaches based on random forests, deep neural networks, and convolutional neural networks have demonstrated improved performance relative to traditional algorithms~\cite{su2020refining,zbizika2022deep}, most retain a pixel-wise formulation or incorporate only limited spatial context. As a result, long-range spatial dependencies that reflect the coherent structure of atmospheric aerosol fields remain underutilized~\cite{jiang2023aerosol}.

The launch of NASA’s Plankton, Aerosol, Cloud, ocean Ecosystem (PACE) mission provides new opportunities for AOD retrieval through hyperspectral observations from the Ocean Color Instrument (OCI), covering ultraviolet to shortwave infrared wavelengths~\cite{nasa_pace_oci_2024}. While hyperspectral data offer enhanced sensitivity to aerosol properties, they also introduce significant spectral redundancy that challenges conventional retrieval strategies~\cite{liang2022estimation}. In parallel, Foundation Models~\cite{jakubik2023foundation,li2025hyperfree,wang2025hypersigma,braham2025spectralearth,tushar2026hyperfm} based on self-attention mechanisms have shown strong performance in remote sensing tasks requiring spatial reasoning, motivating their application to atmospheric retrieval problems~\cite{braham2025spectralearth,tushar2026hyperfm}.

\textbf{In this work, we investigate the use of Foundation AI Models for spatially aware AOD estimation and introduce ViTCG, a novel compact hyperspectral-based vision transformer model with channel-wise grouping}. ViTCG, inspired from HyperFM~\cite{tushar2026hyperfm} foundation model, explicitly models long-range spatial dependencies while efficiently handling high-dimensional spectral inputs. Experimental results demonstrate that ViTCG reduces mean squared error by 62\% relative to a multispectral foundation model baseline and by at least 63\% compared to existing hyperspectral learning-based approaches. Validation against collocated AERONET observations further confirms improved agreement and reduced retrieval error, highlighting the potential of spatially informed foundation models for high-resolution AOD mapping.

\section{Methodology}

\textbf{Problem Formulation.} We formulate Aerosol Optical Depth (AOD) estimation from satellite radiance measurements as a spatial regression problem, rather than as independent pixel-wise retrievals. The proposed model operates on a three-dimensional input tensor of size $C \times H \times W$, where $C$ denotes the number of spectral channels and $H$ and $W$ represent the spatial dimensions of the region. The model outputs a spatially continuous AOD field of size $ H \times W$. This formulation enables the joint exploitation of spectral information and spatial context, allowing coherent AOD patterns to be inferred across extended regions within a single forward pass. As a result, the approach improves computational efficiency while better capturing the spatial structure inherent in atmospheric aerosol distributions. 

\begin{figure}[hbt]
	\centering
	\includegraphics[width=\linewidth]{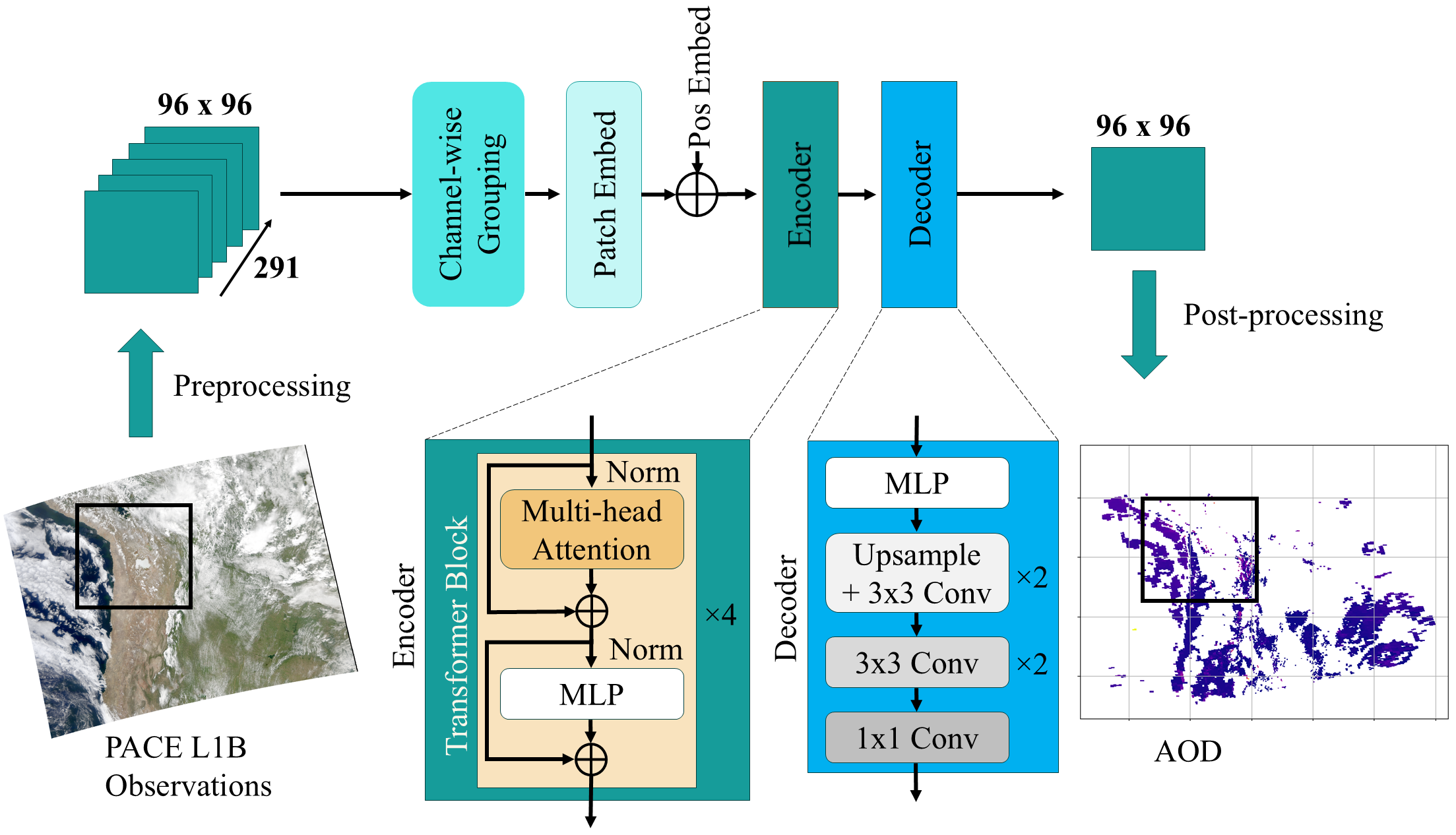}
	\caption{Our Proposed ViTCG model for AOD Prediction }\label{fig:arch}
\end{figure}

\textbf{Foundation AI models for AOD estimation:} 
Foundation AI models provide a natural framework for AOD estimation by capturing nonlocal spatial dependencies and leveraging contextual information across large spatial extents. Atmospheric aerosol fields exhibit strong spatial coherence driven by transport processes and meteorological conditions that are not fully exploited by traditional pixel-wise retrieval methods. Self-attention mechanisms enable adaptive aggregation of information from spatially distant yet physically related regions, improving robustness over heterogeneous surfaces and noisy observations. Through large-scale pretraining, foundation models learn transferable spectral-spatial representations that can be efficiently adapted to aerosol retrieval tasks via fine-tuning, complementing physically based retrieval principles by providing spatial regularization where physical constraints alone are insufficient. Recent multispectral and hyperspectral foundation models, such as PrithviEO1~\cite{jakubik2023foundation}, HyperFree~\cite{li2025hyperfree}, HyperSigma~\cite{wang2025hypersigma}, and SpectralEarth~\cite{braham2025spectralearth}, have demonstrated strong performance in remote sensing applications including land cover mapping and cloud property retrieval; however, to the best of our knowledge, they have not been explicitly designed or systematically evaluated for AOD estimation. Motivated by these observations, we design \textbf{ViTCG} to adapt foundation model principles for spatially coherent and physically consistent aerosol retrieval.

\textit{Our Proposed Model}: We introduce \textbf{ViTCG: Vision Transformer with Channel-wise Grouping (ViTCG)}, a compact vision transformer architecture designed for hyperspectral Aerosol Optical Depth (AOD) estimation from PACE radiance observations. ViTCG incorporates a \textit{channel-wise grouping} strategy to efficiently process high-dimensional spectral inputs while preserving spatial information. By grouping spectrally adjacent channels, the model exploits the strong spectral redundancy inherent in hyperspectral radiance data, reducing parameter count and computational cost without sacrificing retrieval accuracy. An overview of the architecture is shown in Fig.~\ref{fig:arch}. To further limit model complexity, the encoder consists of four transformer blocks, substantially fewer than the twelve blocks used in the standard ViT-Base architecture~\cite{dosovitskiy2020image}. The decoder is lightweight and comprises a sequence of upsampling and convolutional layers to recover full-resolution AOD fields. To ensure fair comparison across different foundation models, an identical decoder configuration is used in all experiments.

\textit{Channel-wise Grouping:}
Channel-wise grouping is introduced to address the strong spectral redundancy inherent in hyperspectral radiance observations and to mitigate excessive information compression at the patch embedding stage. Rather than embedding the full spectral dimension directly, spectrally adjacent channels are grouped prior to patch formation, enabling early-stage spectral mixing while preserving spatial structure. This design allows the model to learn joint spectral-spatial representations within each group while reducing the dimensionality of subsequent embeddings. Formally, the input radiance tensor $\mathbf{X} \in \mathbb{R}^{C \times H \times W}$ is reshaped into a grouped representation $\mathbf{X}_g \in \mathbb{R}^{G \times \frac{C}{G} \times H \times W}$, where $G$ denotes the number of channel groups and each group contains $\frac{C}{G}$ spectral channels.

\textit{Patch and Positional Embedding:}
Following channel-wise grouping, each grouped tensor of size $\mathbb{R}^{\frac{C}{G} \times H \times W}$ is independently partitioned into fixed-size patches and projected into a latent embedding space using a patch embedding layer [token dim $768$]. Positional embeddings are added to each patch embedding to encode spatial location information. The resulting token sequences are then passed to the transformer encoder, as illustrated in Fig.~\ref{fig:arch}.

\textit{Encoder:}
The encoder consists of four transformer blocks. Each block includes a multi-head self-attention (MHSA) layer followed by a feed-forward multilayer perceptron (MLP), with residual connections and layer normalization applied after each sub-layer. The MHSA mechanism enables the modeling of long-range dependencies across spatial locations and spectral groups, while the MLP provides nonlinear feature transformation. This structure supports efficient global context aggregation and progressive feature refinement for AOD estimation. 

\textit{Decoder:}
The decoder is lightweight and comprises an MLP followed by a sequence of upsampling and convolutional layers, as shown in Fig.~\ref{fig:arch}. Encoder tokens are first projected through the MLP and then progressively upsampled to recover the full spatial resolution of the AOD field. Convolutional layers are applied to refine spatial details and improve reconstruction quality. Normalization and activation layers follow each sub-layer but are omitted from the figure for clarity.

Overall, ViTCG follows the standard vision transformer formulation while incorporating channel-wise grouping to better preserve spatial-spectral information in hyperspectral inputs, leading to improved AOD retrieval performance. Our model (as well as other foundation AOD retrieval models) is trained using a mean squared error (L2) loss between predicted and reference AOD values. Invalid or missing pixels are masked out during loss computation to ensure that only valid observations contribute to model optimization.

\section{Experimental Settings}

\textbf{Data:}
We utilize hyperspectral radiance and AOD products from the PACE mission’s Ocean Color Instrument (OCI). The dataset comprises Level-1B (L1B) top-of-atmosphere (TOA) radiance observations spanning 291 spectral bands and corresponding Level-2 (L2) AOD products. Global data granules from 2024 and 2025 are randomly sampled to ensure broad seasonal and geographic coverage. Data preprocessing consists of five steps: (1) filtering invalid pixels using instrument quality flags; (2) geospatial and temporal alignment of L1B and L2 granules; (3) selecting AOD at 550~nm and resampling the 8.4~km L2 AOD products to 1.2~km resolution via interpolation to match the L1B grid; (4) extracting non-overlapping spatial patches of size $96 \times 96$ pixels; and (5) discarding patches containing more than 1\% invalid pixels. To prevent data leakage, training, validation, and test splits are performed at the granule level prior to patch extraction.
Each AOD retrieval model takes as input an L1B radiance patch of size $\mathbb{R}^{291 \times 96 \times 96}$ and predicts a corresponding AOD patch of size $\mathbb{R}^{96 \times 96}$. A total of 1,800 patches are used for training, 250 for validation, and 2,000 held-out patches for testing.

\textbf{Model Comparisons:}
ViTCG is evaluated against a diverse set of baseline models, including a pixel-wise one-dimensional deep neural network (1D DNN)~\cite{yeom2021estimation}, the multispectral foundation model PrithviEO1~\cite{jakubik2023foundation}, and three state-of-the-art hyperspectral foundation models: HyperFree~\cite{li2025hyperfree}, HyperSigma~\cite{wang2025hypersigma}, and SpectralEarth~\cite{braham2025spectralearth}. Convolutional neural network (CNN)-based approaches~\cite{si2024enhanced, jiang2023aerosol} are not included, as they typically require large spatial contexts (e.g., $128 \times 128$ pixels) to retrieve a single output pixel, resulting in substantially higher computational cost.

To establish a rigorous pixel-wise baseline, we evaluate two 1D DNN variants using radiance inputs with 8 spectral bands (DNN\_8w) and the full 291-band spectrum (DNN\_291w). For PrithviEO1, we select the six PACE bands that most closely correspond to its original training wavelengths. For the hyperspectral foundation models, the full PACE spectrum is utilized, with linear interpolation applied within their patch embedding layers to ensure spectral compatibility.

\textbf{Evaluation Metrics:}
Model performance is assessed using multiple complementary statistical metrics, including mean squared error (MSE), root mean squared error (RMSE), mean bias error (MBE), and the index of agreement (IOA)~\cite{yeom2021estimation}. These metrics collectively quantify error magnitude, predictive accuracy, bias, and agreement between retrieved AOD and reference products. In addition to comparisons with PACE L2 AOD, selected results are evaluated against ground-based AERONET observations to assess retrieval consistency with independent measurements~\cite{holben1998aeronet}.

\textbf{Implementation Details:}
All models are implemented in Python using the PyTorch framework. We evaluate a range of hyperparameters, including learning rate, batch size, learning-rate schedulers, and optimization algorithms. ViTCG achieves optimal performance with a learning rate of $10^{-4}$ and an effective batch size of 256. Model optimization is performed using the AdamW optimizer, and early stopping is applied based on validation MSE with a patience of 50 epochs.

\section{Results and Discussion}

\textbf{Quantitative Results.} 
Table~\ref{tab:results} summarizes the quantitative performance of all evaluated models for AOD estimation. ViTCG outperforms all compared AOD methods, achieving at least a 62\% reduction in mean squared error (MSE), a 38\% reduction in root mean squared error (RMSE), and the lowest mean bias error. In addition, ViTCG attains the highest index of agreement (IOA) among all models, indicating improved consistency with the reference AOD products. The results demonstrate that channel-wise grouping enhances the extraction of informative spectral-spatial features from hyperspectral data. Models without channel-wise grouping exhibit increased bias and error, highlighting the importance of this design choice. Furthermore, despite access to the full spectral input, the limited capacity of the pixel-wise DNN baseline restricts its ability to effectively exploit spectral redundancy, resulting in inferior performance.
\begin{table}[ht]
    \centering
    \caption{Comparison of AOD Retrievals Methods.}
    \label{tab:results}

    \resizebox{\columnwidth}{!}{%
    \begin{tabular}{r c c c c}
        \toprule
\textbf{\begin{tabular}[c]{@{}c@{}}Retrieval \\ Methods\end{tabular}} & 
\textbf{MSE  ($\downarrow$)} & 
\textbf{RMSE ($\downarrow$)} & 
\textbf{MBE  ($\downarrow$)} & 
\textbf{IOA  ($\uparrow$)} \\
        \cmidrule(lr){1-1} \cmidrule(lr){2-2}\cmidrule(lr){3-3}\cmidrule(lr){4-4}\cmidrule(lr){5-5}
        DNN\_8w~\cite{yeom2021estimation} &$0.0441$ & $0.2099$ & $-0.1190$ & $0.3496$ \\ 
        DNN\_291w~\cite{yeom2021estimation} &$0.0426$ & $0.2063$ & $-0.1123$ & $0.3504$ \\ 
        PrithviEO1~\cite{jakubik2023foundation}  & 
        $0.0370$ & $0.1925$ & $-0.0838$ & $0.3721$ \\
        
        HyperFree~\cite{li2025hyperfree}  & 
        $0.0516$ & $0.2272$ & $-0.0889$ & $0.3478$ \\
        
        HyperSigma~\cite{wang2025hypersigma}  & 
        $0.0385$ & $0.1963$ & $-0.0811$ & $0.3626$ \\
        
        SpectralEarth~\cite{braham2025spectralearth}  & 
        $0.0453$ & $0.2129$ & $-0.0802$ & $0.4025$ \\
        
        ViTCG [no CG]  & 
        $0.0227$ & $0.1506$ & $0.0498$ & $0.7603$ \\
 
        \textbf{ViTCG (ours)}  & 
        $\mathbf{0.0141}$ & 
        $\mathbf{0.1186}$ & 
        $\mathbf{0.0008}$ & 
        $\mathbf{0.8552}$ \\
        \bottomrule
    \end{tabular}
    }
\end{table}

\begin{figure}[ht]
	\centering

    \includegraphics[width=\linewidth]{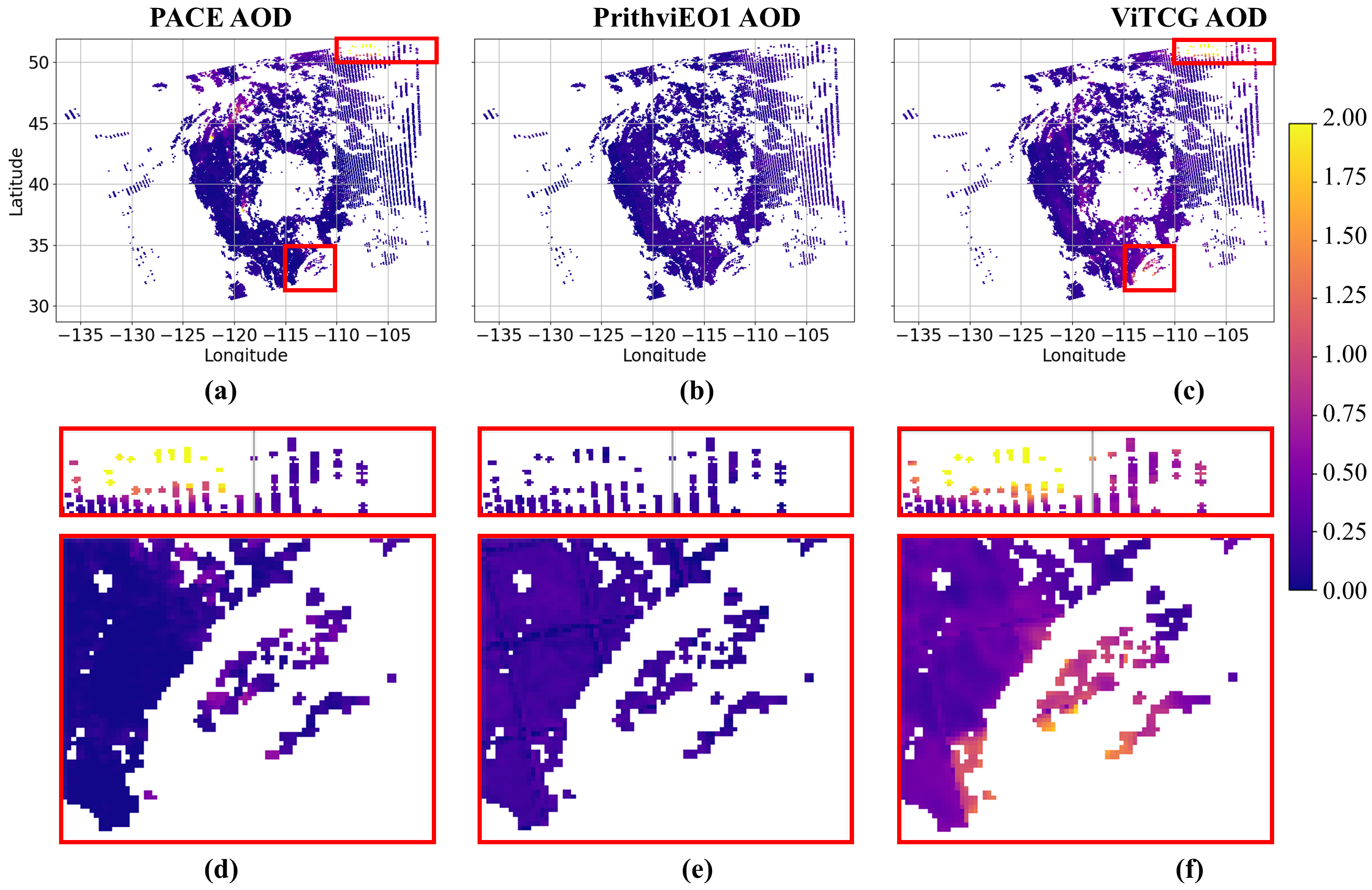} 

        \caption{AOD Estimation from PACE-OCI observation on September 1st 2025 (UTC 20:15:49) over USA. Top row: AOD from PACE, PrithviEO1 and ViTCG respectively; Bottom row: Highlighted regions.}
    \label{fig:full_granule}
\end{figure}

\textbf{Qualitative Results.} 
Qualitative comparisons of AOD retrievals are shown in Fig.~\ref{fig:full_granule}. A region over the United States is randomly selected from the PACE OCI L1B product, ensuring that both the spatial extent and acquisition time are disjoint from the training and validation datasets. Retrieval is performed using a sliding-window approach, where a $96 \times 96$ window is moved across the scene to estimate AOD over the full region.
The results show that \textit{PrithviEO1}, a foundation model, avoids high AOD ranges, exhibits boundary artifacts and block-like patterns near patch edges, indicating sensitivity to window-based inference. In contrast, the ViTCG produces spatially uniform AOD retrievals. Although isolated failures occur [Fig.~\ref{fig:full_granule}f], the overall output exhibits superior spatial coherence and structural robustness across the entire domain.

\begin{figure}[h]
	\centering
	\includegraphics[width=\linewidth]{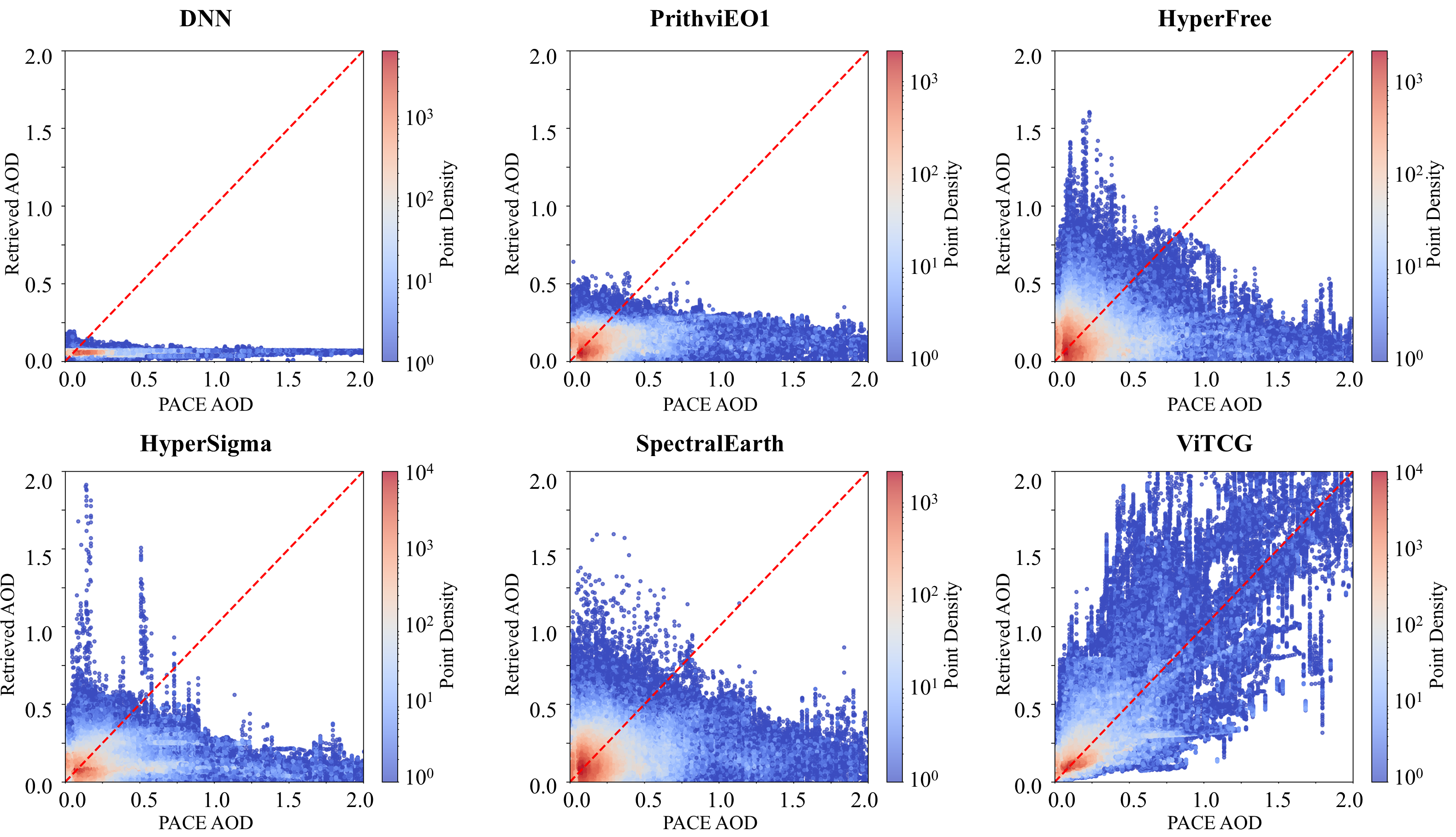}
    \caption{Comparison between PACE AOD and retrieved AOD using different methods on held-out test patches.}
    \label{fig:scatter}
\end{figure}

Fig.~\ref{fig:scatter} shows that ViTCG most accurately captures the full distribution of AOD values. In contrast, the pixel-wise DNN is confined to a narrow dynamic range, while PrithviEO1 exhibits conservative saturation at AOD values around $0.5$. Although hyperspectral foundation models cover a broader range, they systematically underrepresent high-AOD events.

\textbf{Computing Comparison.} 
All models are evaluated under comparable computational settings, with inference time measured on a single 24~GB RTX~6000 GPU and training performed on the GPUs specified for each method. ViTCG contains 8.82~M parameters, approximately $10\times$ fewer than multispectral and hyperspectral foundation model baselines (87–102~M), resulting in substantially lower computational cost. ViTCG trains in 19:45 hours on four 12~GB RTX~2080~Ti GPUs and achieves an inference time of 0.0069~s per $96 \times 96$ patch, compared to 30–57 hours of training and 0.0076–0.0309~s inference time for existing foundation models on four 24~GB RTX~6000 GPUs. Although pixel-wise 1D DNN models have low per-pixel latency (0.0005~s), they scale poorly; processing a $96 \times 96$ region requires 4.608~s, approximately $668\times$ slower than ViTCG. By jointly modeling spatial–spectral patches, ViTCG enables efficient large-area AOD retrieval that is impractical for pixel-wise or high-parameter foundation models.

\begin{figure}[h]
	\centering
   	\includegraphics[width=0.9\linewidth]{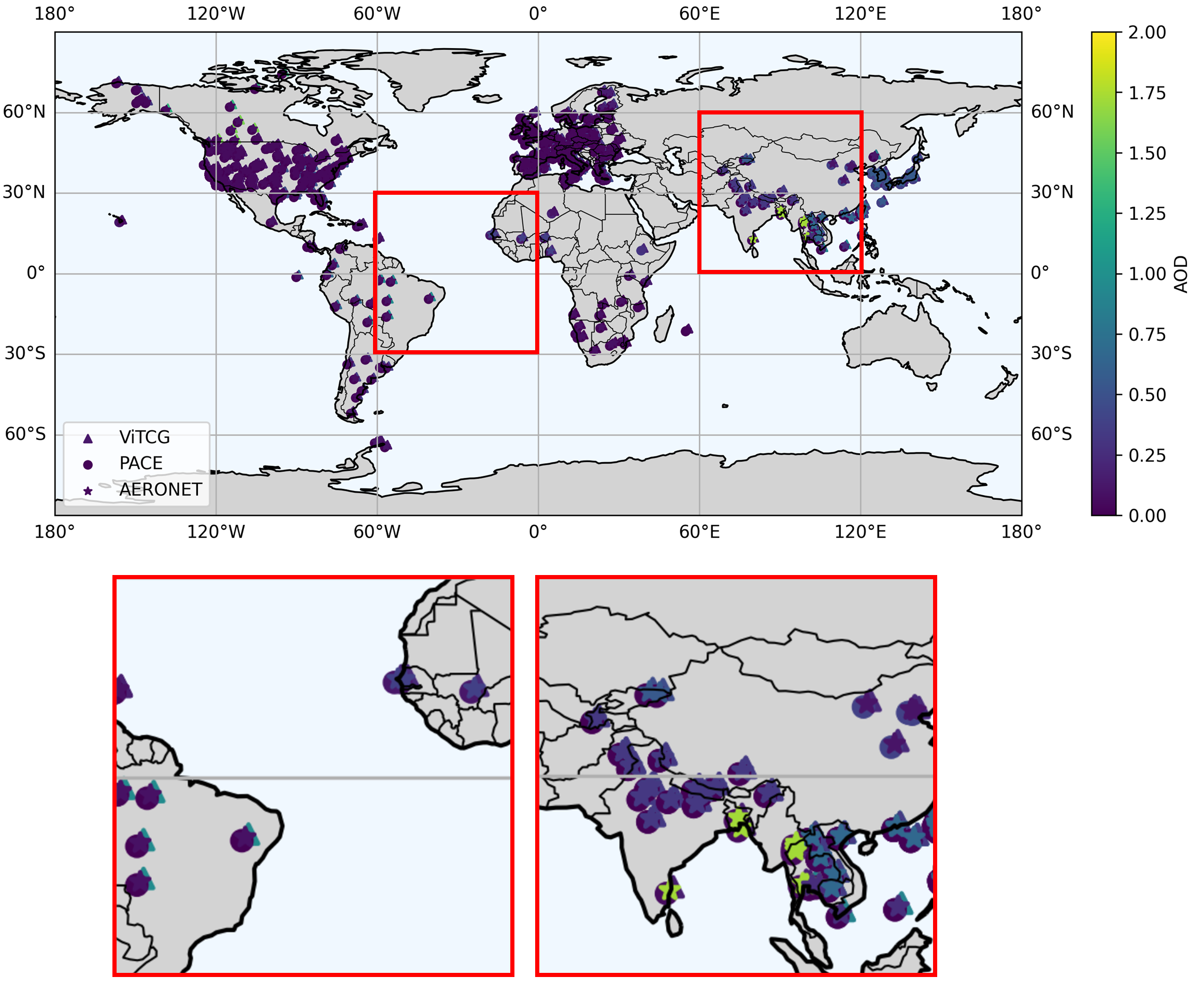} 
	\caption{Global Validation of ViTCG AOD against AERONET.}\label{fig:world}
\end{figure}

\textbf{Validation against AERONET:}
AERONET (AErosol RObotic NETwork)~\cite{dubovik2000accuracy} is widely regarded as the reference standard for validating satellite-derived cloud and AOD products due to its highly accurate ground-based sun photometer measurements~\cite{she2020himawari, yeom2021estimation, liang2022estimation}. However, AERONET provides point-based observations, whereas satellite retrievals such as PACE represent spatial averages over finite footprints (e.g., $8.4~\mathrm{km} \times 8.4~\mathrm{km}$), leading to inherent representativeness differences.
For validation, AERONET Level~2 AOD measurements at 500~nm were temporally matched within $\pm30$ minutes of PACE overpasses and spectrally adjusted to 550~nm using the \r{A}ngstr\"om exponent~\cite{yeom2021estimation}. The analysis focuses on 24-hour periods on February~20 and March~20, 2025, during which PACE OCI L1B radiance and L2 AOD granules were selected based on AERONET site locations and processed following the Experimental Settings. Non-overlapping $96 \times 96$ patches centered on AERONET sites were extracted, yielding 125 spatiotemporally collocated samples after quality filtering. ViTCG predictions were averaged over a central $8 \times 8$ pixel region to maintain spatial consistency with the satellite footprint. As shown in Fig.~\ref{fig:world}, ViTCG demonstrates strong global agreement with AERONET observations, with larger deviations in high-AOD regions such as South America and the Indian subcontinent, primarily attributable to spatial representativeness error arising from localized aerosol plumes.
\vspace{-10pt}
\section{Conclusions}
Accurate estimation of Aerosol Optical Depth (AOD) is essential for air quality monitoring and climate research. We present ViTCG, a compact spatial regression framework that overcomes key computational limitations of pixel-wise AOD retrieval methods by capturing spatial-spectral coherence in PACE OCI hyperspectral observations with an order-of-magnitude reduction in parameters. Experimental results demonstrate that ViTCG consistently outperforms both multispectral and hyperspectral foundation model baselines, highlighting its potential for scalable, high-resolution aerosol monitoring in future hyperspectral Earth observation missions. Future work will focus on extending ViTCG into a foundation model that incorporates physical constraints and generalizes across longer time periods and diverse aerosol regimes.

\hspace{3pt}\textbf{Acknowledgment:}
This research is partially supported by NSF grant 2238743, and carried out using the computational facilities of the High Performance Computing Facility, UMBC.

\small
\bibliographystyle{IEEEtranN}
\bibliography{references}

\end{document}